\documentclass[letterpaper, 10 pt, conference]{ieeeconf}  

\IEEEoverridecommandlockouts                              

\usepackage{graphics} 
\usepackage{epsfig} 
\usepackage{mathptmx} 
\usepackage{times} 
\usepackage{amsmath} 
\usepackage{amssymb}  

\usepackage{url}
\usepackage{booktabs} 
\usepackage{subfigure}

\title{\LARGE \bf
Emotional Tandem Robots: How Different Robot Behaviors Affect Human Perception While Controlling a Mobile Robot
}

\author{Julian Kaduk$^{1}$, Friederike Weilbeer$^{2}$ and Heiko Hamann$^{1}$
\thanks{*This work was supported by European Union’s Horizon 2020 FET Open research program, grant no 964464, project ChronoPilot and by Deutsche Forschungsgem., Germany's Excellence Strategy EXC 2117-422037984.}
\thanks{$^{1}$Julian Kaduk and Heiko Hamann are with the Department of Computer and Information Science,
        University of Konstanz, 78457 Konstanz, Germany
        {\tt\small julian.kaduk@uni-konstanz.de},
        {\tt\small heiko.hamann@uni-konstanz.de}}%
\thanks{$^{2}$Friederike Weilbeer is with the Department of Computer and Information Science, University of Lübeck,
        23562 Lübeck, Germany
        }
}

\begin{document}

\maketitle
\thispagestyle{empty}
\pagestyle{empty}

\begin{abstract}

In human-robot interaction (HRI), we study how humans interact with robots, but also the effects of robot behavior on human perception and well-being. Especially, the influence on humans by tandem robots with one human controlled and one autonomous robot or even semi-autonomous multi-robot systems is not yet fully understood. Here, we focus on a leader-follower scenario and study how emotionally expressive motion patterns of a small, mobile follower robot affect the perception of a human operator controlling the leading robot. We examined three distinct emotional behaviors for the follower compared to a \emph{neutral} condition: \emph{angry}, \emph{happy} and \emph{sad}. We analyzed how participants maneuvered the leader robot along a set path while experiencing each follower behavior in a randomized order. We identified a significant shift in attention toward the follower with emotionally expressive behaviors compared to the \emph{neutral} condition. For example, the \emph{angry} behavior significantly heightened participant stress levels and was considered the least preferred behavior. The \emph{happy} behavior was the most preferred and associated with increased excitement by the participants. Integrating the proposed behaviors in robots can profoundly influence the human operator's attention, emotional state, and overall experience. These insights are valuable for future HRI tandem robot designs. 

\end{abstract}

\section{Introduction}\label{section:intro}   
We aim to enhance interactions with robots to be more intuitive and natural in HRI. Research has shown the importance of transmitting emotions through verbal and non-verbal cues, including facial expressions, vocal aspects, and body motion~\cite{Dael.2012}. Humanoid robots can mimic these cues, but non-humanoid (mobile) robots typically lack such capabilities due to their limited degrees of freedom, different shapes, and sizes. Yet, non-humanoid robots (e.g., vacuum cleaners) become more prevalent in our daily lives and are commonly anthropomorphized~\cite{Hendriks.2011} as their human owners attribute them human-like characteristics or give them names. Studies suggest that designing specific personalities for these robots could improve user experience~\cite{Hendriks.2011}.

Communicating emotions will be crucial as we move from autonomous service robots to interactive human-robot collaboration, which is becoming more feasible and increasingly popular~\cite{Natarajan.2023}. Humans are able to reason in complex ways to solve diverse and unique tasks. Robots, in turn, can contribute super-human capabilities that exceed the size, strength, and speed of humans while operating in hostile environments or repeating monotonous tasks for extended duration. Scaling this idea to multi-robot systems and robot swarms~\cite{hamann2018}, the collaboration provides added potential by benefiting from the high number of rather simple robots and distributed control~\cite{Bashyal.2008knf}. As the swarm behavior emerges through local interactions, they autonomously adapt to complex environments and dynamic missions. However, human high-level oversight is still required to supervise and guide robot swarms in real-world applications. 

Controlling two or many robots requires specially designed interfaces because humans cannot divide their attention to multiple targets simultaneously for individual control of each robot~\cite{pendelton.2013}. A~method to overcome this limitation in swarm robotics is the leader-follower approach, where one or a subset of robots is selected to be user-controlled, guiding the behavior of others~\cite{Kolling.2016,goodrich2012leadership}. Based on inspiration from natural collective behavior, this approach realizes distributed control within the multi-robot system or robot swarm while allowing control from a human operator with access to global information. Local interactions and simple behavioral rules between robots facilitate scalability, flexibility, and robustness, making it suitable for a wide range of applications~\cite{hamann2018}. However, large robot swarms require a novel robot-human communication paradigm because transmitting status information about each robot is infeasible~\cite{divbandSoorati21}. Instead, expressive behavior could be leveraged for communication.

Recent work showed that emotions can be expressed though the motion of a swarm of miniature mobile robots and that the emotion is recognized and classified correctly by human observers~\cite{Santos.2021}. Further research underscores changes in the speed and smoothness of robot motion having an effect on the perceived emotion of robots in a swarm~\cite{Dietz.2017}. The same study also showed that smoothness affected the participants' emotional valance and speed the participants' emotional arousal levels. However, these results were obtained in a passive interaction scenario. To the best of our knowledge, the question whether these types of expressive motion patterns affect human perception in an active interaction task is open. Here, we study effects of emotionally expressive robot motion on an active human collaborator. We chose the leader-follower approach to lay the groundwork for future investigations in the direction of human-swarm interaction (HSI) but limited the scenario to two small mobile robots: one follower robot that follows one human-controlled leader robot. Our rationale is to reduce experiment complexity and to isolate the effect of expressive motion from possible magnitude or size effects with a robot swarm. 

We investigated the effects on the human operator of three emotional behaviors (\emph{angry}, \emph{happy} and \emph{sad}) compared to a \emph{neutral} control condition. The results allow us to better understand their influence on human-robot dynamics and potential applications in HRI. Future studies are needed to research the nuanced impacts of robotic emotions across different demographics and in varied contexts. Our findings give insights about how the operator's attention can be guided and their emotional state be effected. We provide an overview of related work on emotionally expressive behavior in human robot interaction and how it is linked to psychological effects on human perception. We provide a detailed description of our study design and present our results.

\section{Related work}\label{section:literature} 
The human capacity to communicate nonverbal emotions serves as a foundation for our daily interactions. Research on this ability has a long history and is fundamental for social communication~\cite{Shariff.2011}. Nowadays, with the rapid integration of robots into our daily lives, the interest and necessity to empower them with better expressive capabilities are growing. For many years, it has been studied how robots can express emotions and how humans can perceive them and react to them. For example, see the recent survey by Stock-Homburg on humanoid robots~\cite{Stock-Homburg.2022}.

Humanoid robots designed for direct social interactions to replace humans require the ability to establish an emotional connection with their human counterparts beyond simple information transfer. Crucial examples of this necessity are the use of robots in education~\cite{newton2019humanoid} or healthcare~\cite{Pepito.2020}. As a result, they are often designed to mimic human features, which can be used to express emotions similarly to humans. Although some robots, such as Nao, can use their whole body to present emotional postures~\cite{Erden.2013}, others are limited to facial features~\cite{Berns.2006}. The human face and its complex movements have long been studied as a modality to express our emotions~\cite{ekman1979facial}. In robotics, facial actuation comes with challenges. For example, the humanoid robot Kismet, has 15~degrees of freedom dedicated to move its facial features~\cite{Breazeal.2003}. 

Non-humanoid robots are not designed for that and lack actuation for complex expressive features. However, there is a small body of literature on emotionally expressive non-humanoid robots~\cite{Wang.2021tv}. Examples range from simple projections of abstract geometric shapes moving on a wall~\cite{Mutlu.2006}, to abstract robots in the form of a wood shaft able to roll, pitch, and yaw around a single point~\cite{Harris.2011}, to simple wheeled robots that communicate emotions through their motion, color, and sound~\cite{Löffler.2018as}. Although the ability to show artificial emotions is crucial for the social acceptance of robots in public spaces, the implementation of universally recognizable patterns has been shown to be challenging~\cite{Hoggenmueller.2020}.

Our ongoing research, although currently focused on robot groups of two, intends to contribute to swarm robotics. The complexity and expressive potential of swarm systems enhance the human-swarm interaction (HSI) landscape, demanding innovative approaches to communication, such as a "joint human-swarm loop" that integrates robots with the human's neural activities~\cite{Hasbach.2022}. This integration might involve robots adapting to human emotional states, inferred from physiological signals~\cite{Villani.2020cve}.

Research on conveying emotions through the movement of robots is growing. Studies have illustrated how varying the synchronicity and proximity of robots within a swarm can effectively convey emotions like fear, happiness, and surprise~\cite{St-Onge.2019cwd}. Moreover, distinct emotional states have been successfully identified from choreographed robotic movements~\cite{Santos.2021} and it has been shown that the nature of robot motion (smooth vs. jittery) and speed can influence observers' emotional responses~\cite{Dietz.2017}. Further research has illustrated how the robot speed and the number of robots can influence the time perception, perception of flow, cognitive demand~\cite{Kaduk.2023} and the emotional valence and arousal of a human operator~\cite{Cavdan.2023}. 

While these investigations shed light on how robots can expressively communicate and that robot motion has an effect on human perception, the impact on humans actively interacting with the emotionally expressive robots remains underexplored.

\section{Method}\label{section:method}   

\subsection{Scenario}
We developed a system in which a user-controlled mobile leader robot is followed by an autonomous mobile follower robot. The follower was designed to emulate one of three potentially distinct emotionally expressive behaviors: \emph{angry}, \emph{sad}, and \emph{happy}. We also implemented \emph{neutral} follower behavior to serve as our baseline and control condition.

\begin{figure}[t]
    \centering
    \vspace{8pt}
    \includegraphics[height=6cm]{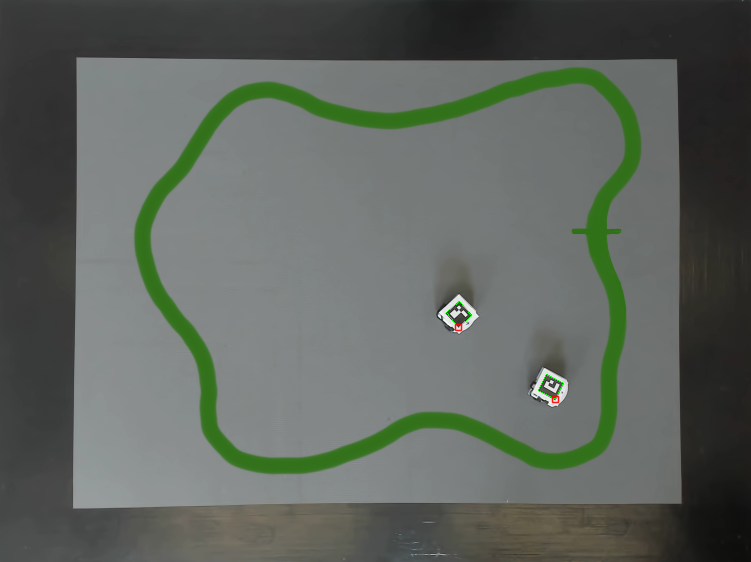}
    \caption{Arena setup for the research study. Path marked in green.}
    \label{fig:studysetup}    
\end{figure}

In our user study, participants were assigned to maneuver the leader robot along a green tape-lined path on the floor, as shown in Fig.~\ref{fig:studysetup}. A~full lap took on average 150 seconds with a standard deviation of 5.42 seconds. The participants repeated this task for all four behaviors. Each trial began with the robots at a designated starting position and ended when the leader finished the lap.

\subsection{Experiment procedure}
The participants received an overview of the research project and the experiment procedure, with the post-trial questionnaire introduced before the first trial to minimize bias. Before the experiment, we briefed them on the leader's controls, and they performed a practice round without the follower robot to familiarize them with the interface. The trials for each behavior were completed in random order. After each trial, the participants filled out the post-trial questionnaire consisting of nine items: a duration estimation task, followed by seven items of 5-point likert-type, and a final question to classify the robot behavior as one of the fundamental emotions (disgust, surprise, happiness, sadness, anger, fear). All questions are shown in Table~\ref{tab:questionnaire}. 

\begin{table*}[t]
  \caption{Questionnaire items}
  \label{tab:questionnaire}
  \centering
    \scriptsize
  \begin{tabular}{l l l}
  \toprule
    \textbf{Question} & \textbf{Scale} & \textbf{Type} \\
    \midrule
    Please estimate the duration of the previous task. & mm:ss & free input \\
    How quickly did time pass for you during the previous task? & very slow - very fast  & 5-pt likert\\
    Did you feel that the follower was keeping up with the leader? & very - not at all & 5-pt likert\\
    How would you rate your distribution of attention on the two robots? & leader - follower & 5-pt likert\\
    How did you feel emotionally during the task? & happy - sad & 5-pt likert\\
    How did you feel emotionally during the task? & relaxed - stressed & 5-pt likert\\
    How did you feel emotionally during the task? & bored - excited & 5-pt likert\\
    How did you feel during the task? & active - calm & 5-pt likert\\
    Which of the fundamental emotion best describes the behavior depicted? & disgust, surprise, happiness, sadness, anger, fear & selection \\
  \bottomrule
  \end{tabular}
\end{table*}

To aid the interpretation of the self-reported measures we conducted short semi-structured interviews at the end of the complete session for each participant. We asked them which behavior they preferred, how they classified the behaviors, and encouraged them to add further comments about the different behaviors and the overall experiment.

\subsection{Robot hardware and leader robot implementation} 
We used the Thymio II robot~\cite{Riedo_Chevalier_Magnenat_Mondada_2013}, enhanced with a Raspberry Pi~3B for wireless connectivity and a Python interface. The robot has dimensions of 110~mm by 112~mm, is equipped with two motors that form a differential drive, and a maximum speed of 0.2~$\frac{m}{s}$. For the follower robot navigation, we fixed ArUco markers on the Thymio's top. For the leading robot, we developed a control interface using standard computer keyboard arrow keys with nine distinct input combinations for different motions.

\subsection{Follower robot implementation}
The follower robot relies on positional input data obtained by overhead tracking using ArUco markers and a ceiling-mounted 1280$\times$720~px RGB camera. The camera feed is centrally processed by OpenCV-based tracking software~\cite{bradski2000opencv}. Using the leader center, leader orientation, follower center, and follower orientation, the follower calculates a goal point~$P_{\text{goal}}$, 70~px behind the leader, $P_{\text{goal}} = L - 70 * \Vec{o_L}$, where $L$ is the center of the leader and $\Vec{o_L}$ its orientation. The vector $\Vec{d}$ describes the Euclidean distance between the follower center and $P_{\text{goal}}$. The angle~$\theta$ between the heading of the follower~$o_F$ and $\Vec{d}$ is calculated to obtain the robots' orientation toward each other. All distances are mapped in pixels with $3.5$~px$\approx1cm$ on the arena surface. Based on these values, the main part of each follower implementation uses the same logic with individual velocities and thresholds. If the leader stops, the follower stops as well. The remaining differences between the behaviors are built on that. All follower robot implementations are shown in the supplementary video.

\subsubsection{Neutral}
The \emph{neutral} follower switches between two states. The first state is activated when distance~$\lVert \Vec{d} \rVert_2<80$~px. At this point the follower stops and waits for the leader to continue moving. The second state is triggered when $\lVert \Vec{d} \rVert_2>80$~px. In this state, the speed of the left and right wheels ($v_l ,v_r$) depends on the value of $\theta$. If $\theta\in[-15$\textdegree$,15$\textdegree$]$, the robot drives forward with ($v_l=v_r=0.1~\frac{m}{s}$). For $\theta\in[-180$\textdegree$,-15$\textdegree$]$, the robot turns to the left, with ($v_l=-0.04~\frac{m}{s}$) and ($v_r=0.04~\frac{m}{s}$). In contrast, when $\theta\in[15$\textdegree$,180$\textdegree$]$, the robot turns to the right, with $v_l=0.04~\frac{m}{s}$ and $v_r=-0.04~\frac{m}{s}$.

\subsubsection{Happy}
The \emph{happy} behavior follows a similar approach to the neutral behavior. Instead of following with $v_l=v_r$, the \emph{happy} robot follows at a higher velocity than the leader with a fast-paced lateral oscillating motion alternating between $v_1=0.04~\frac{m}{s}$ and $v_2=0.16~\frac{m}{s}$ for each of the wheels at a frequency of $10~Hz$. Once the follower has caught up with the leader to $\lVert \Vec{d} \rVert_2<70~px$, it does a 360$^{\circ}$ rotation on the spot with $v_l=0.16~\frac{m}{s}$ and $v_r=-0.16~\frac{m}{s}$ before returning to the oscillating motion.

\subsubsection{Angry}
The \emph{angry} follower is implemented by four states and follows the leader with a much shorter distance than other followers by setting $\lVert \Vec{d} \rVert_2 < 35$~px. Once it reaches this point it stops and waits for the leader to proceed. At larger distances, this implementation randomly iterates over three different patterns every five seconds.

In the first pattern, the \emph{angry} follower moves straight for $\theta\in[-15$\textdegree$,15$\textdegree$]$ and switches between a slow speed is $v_r=v_l=0.04~\frac{m}{s}$ and a fast speed $v_r=v_l=0.16~\frac{m}{s}$ at a frequency of $10~Hz$. For $\theta\in[-180$\textdegree$,-15$\textdegree$]$, the robot turns to the left, with $v_r=-v_l=0.06~\frac{m}{s}$. In contrast, when $\theta\in[15$\textdegree$,180$\textdegree$]$, the robot turns to the right with $v_l=-v_r=0.06~\frac{m}{s}$. This pattern resembles a pushing stop and go motion.

In a second pattern, the \emph{angry} follower moves straight for $\theta\in[-10$\textdegree$,10$\textdegree$]$ with $v_r=v_l=0.18~\frac{m}{s}$. For $\theta\notin[-10$\textdegree$,10$\textdegree$]$, it slowly turns to the right at $v_l=-v_r=0.024~\frac{m}{s}$ if $\theta\in[10$\textdegree$,180$\textdegree$]$ and to the left at $v_r=-v_l=0.024~\frac{m}{s}$ if $\theta\in[-180$\textdegree$,-10$\textdegree$]$.

A third pattern moves the robot in a straight line for $\theta\in[-10$\textdegree$,10$\textdegree$]$ with $v_r=v_l=0.14~\frac{m}{s}$. For $\theta\in[10$\textdegree$,180$\textdegree$]$ it turns right at $v_l=-v_r=0.06~\frac{m}{s}$ and left at $v_r=-v_l=0.06~\frac{m}{s}$ for $\theta\in[-180$\textdegree$,-10$\textdegree$]$.

\subsubsection{Sad}
The \emph{sad} behavior is slower than the leader in its main pattern and therefore, once $\lVert \Vec{d} \rVert_2>200~px$ the follower needs to catch up, which is implemented with $v_r=v_l=0.14~\frac{m}{s}$ when $\theta\in[-15$\textdegree$,15$\textdegree$]$ and a right turn at $v_l=-v_r=0.032~\frac{m}{s}$ and a left turn at $v_r=-v_l=0.032~\frac{m}{s}$ for angles outside of these values. Once the robot has caught up to $\lVert \Vec{d} \rVert_2<100~px$, it switches to a sinusoidal following trajectory. 

The sine trajectory is calculated when the follower has caught up with the leader ($\lVert \Vec{d} \rVert_2<100~px$). It is calculated between the points 30~px behind the leader and 10~px in front of the follower with a frequency $f = 2\pi$ and an amplitude of $a = 20~px$. We split this trajectory into five equally spaced points. The robot then follows a simple algorithm to navigate to the next point. When the next point is within $\theta\in[-15$\textdegree$,15$\textdegree$]$ it drives straight with both wheels at $v_{\text{straight}}$. When the point is at an angle of $\theta\in[-15$\textdegree$,15$\textdegree$]$, the robot turns accordingly with one wheel at $v_{\text{straight}}$ and the other wheel at $v_{\text{turn}}$. The values of $v_{\text{straight}}$ and $v_{\text{turn}}$ are scaled based on the Euclidean norm of $\Vec{d}$ between the two robots. At $\lVert \Vec{d} \rVert_2<100$~px~$ \rightarrow v_{\text{straight}}=0.088~\frac{m}{s}~and~v_{\text{turn}}=0.008~\frac{m}{s}$, at $\Vec{d}>100$~px~and~$\Vec{d}<120$~px~$ \rightarrow v_{\text{straight}}=0.08~\frac{m}{s}~and~v_{\text{turn}}=0.012~\frac{m}{s}$ and at $\Vec{d}>120$~px~$ \rightarrow v_{\text{straight}}=0.072~\frac{m}{s}~and~v_{\text{turn}}=0.016~\frac{m}{s}$. This way, the follower slowly increases the distance to the leader. Once it has completed all points it evaluates its distance to the leader. If ($\lVert \Vec{d} \rVert_2<200$~px) it calculates a new sine trajectory and follows it. Otherwise, the follower switches to the catch-up pattern until $\lVert \Vec{d} \rVert_2<100$~px. If $\lVert \Vec{d} \rVert_2$ is less than 80~px, the robot stops completely and waits.

\subsection{Participants}
The study involved 15 participants, with ten identifying as males and five as females, and an average age of 24.5 years. 13 participants were university students and nine of them had previous robotics experience. We had to exclude one participant due to a hardware failure that affected the robot's speed and extended the trial duration.

\section{Results}\label{section:results}  
We analyzed the self-reported measures to investigate the effects of the different robot behaviors on the participants. The results are given as the mean ($M$) and the standard error of the mean ($SE$).

\subsection{Participant perception of the robot}
\begin{figure}[t]
    \subfigure{\includegraphics[width=1.55in]{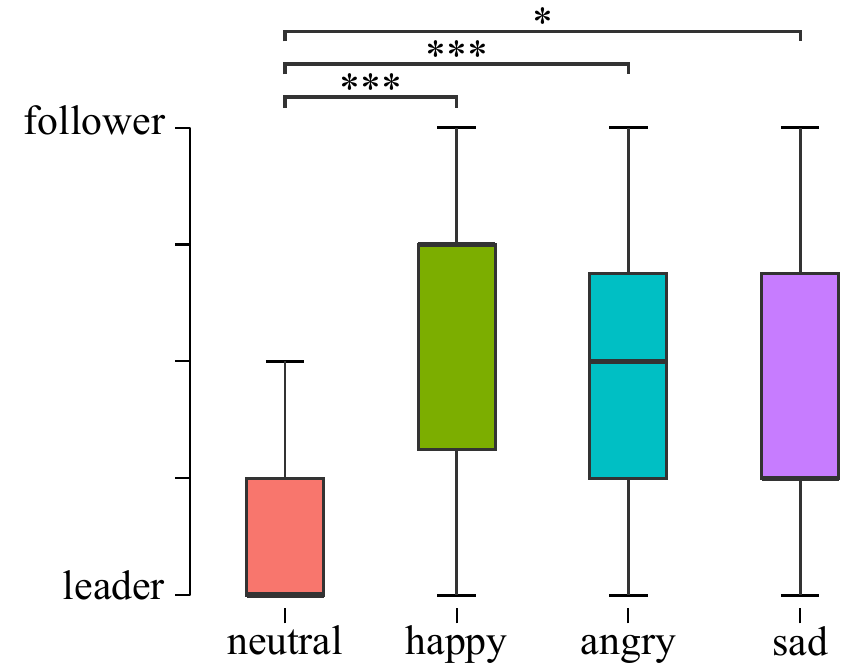}} 
    \subfigure{\includegraphics[width=1.55in]{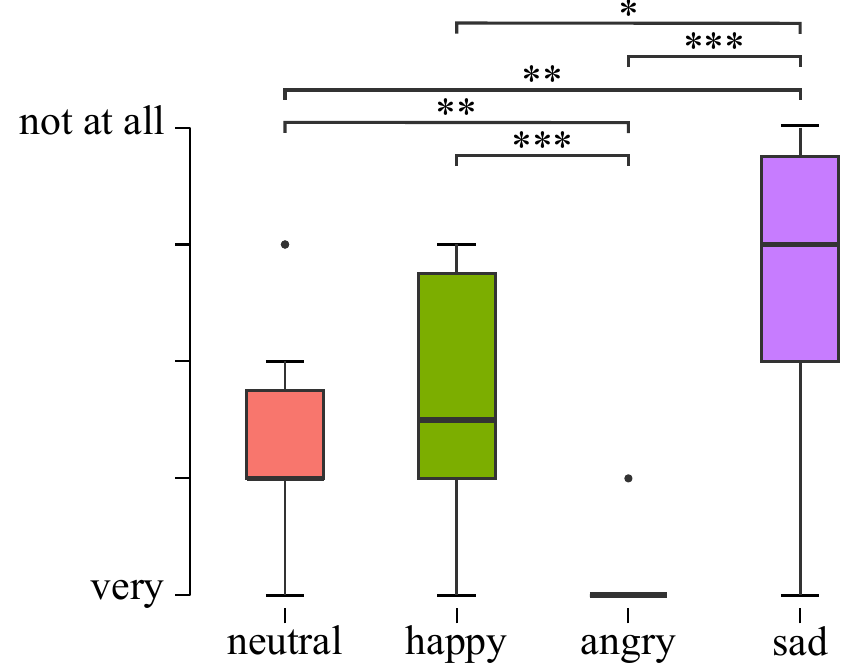}}
    \caption{Attention distribution on the left and follower's perceived capability to keep up on the right. Significant effects are indicated as * for $p_{\text{bonf}}\leq0.05$, ** for $p_{\text{bonf}}\leq0.01$, *** for $p_{\text{bonf}}\leq0.001$.}
    \label{fig:perception}
\end{figure}

An analysis using repeated measures ANOVA whether the follower could keep pace with the leader and which robot attracted more of their attention during the trial revealed a significant main effect for both, $p \leq .001$, as shown in Fig.~\ref{fig:perception}. Large effect sizes support this finding, with $\eta p^2 = .40$ for the attention distribution and $\eta p^2 = .72$ indicating how well the follower kept up.

Subsequent Bonferroni post-hoc tests showed that during the \emph{neutral} behavior, participants' attention significantly shifted towards the leader ($M=1.43$, $SE=.17$) in contrast to the \emph{angry} ($M=2.86$, $SE=.29$), \emph{sad} ($M=2.64$, $SE=.32$) and \emph{ happy} behaviors ($M=3.21$, $SE=.33$). However, there was no difference in attention among the three emotionally expressive behaviors.

Further looking into the Bonferroni post-hoc results for the follower's ability to keep pace, the \emph{angry} behavior ($M=1.08$, $SE=.07$) was perceived to keep up better, especially when compared to both the \emph{happy} ($M=2.69$, $SE=.30$) and \emph{neutral} behaviors ($M=2.31$, $SE=.26$). Conversely, the \emph{sad} behavior ($M=4.00$, $SE=.22$) was perceived to lag behind more than the other behaviors. 

\subsection{Emotion}
\begin{figure}[t]
\vspace{1pt}
    \subfigure{\includegraphics[width=1.55in]{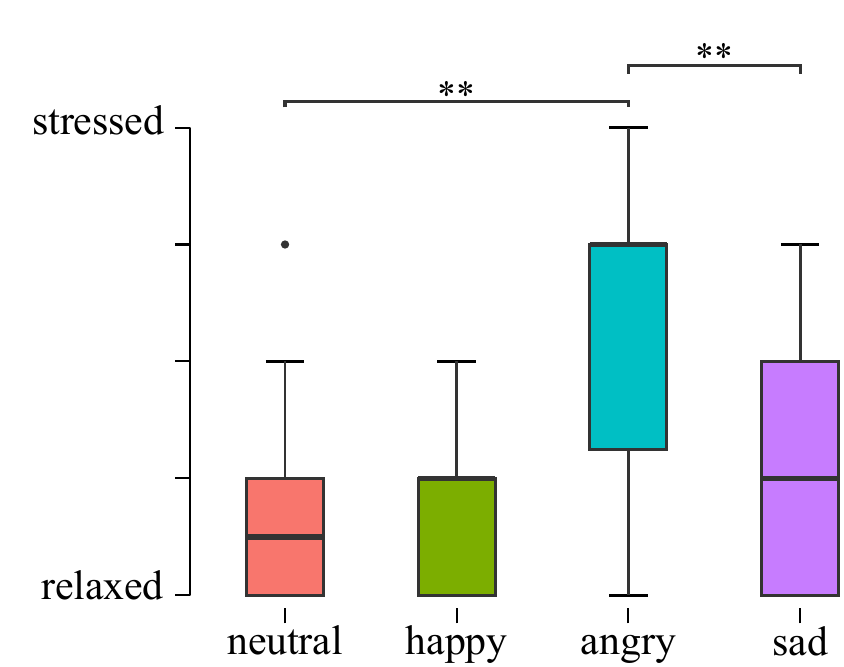}} 
    \subfigure{\includegraphics[width=1.55in]{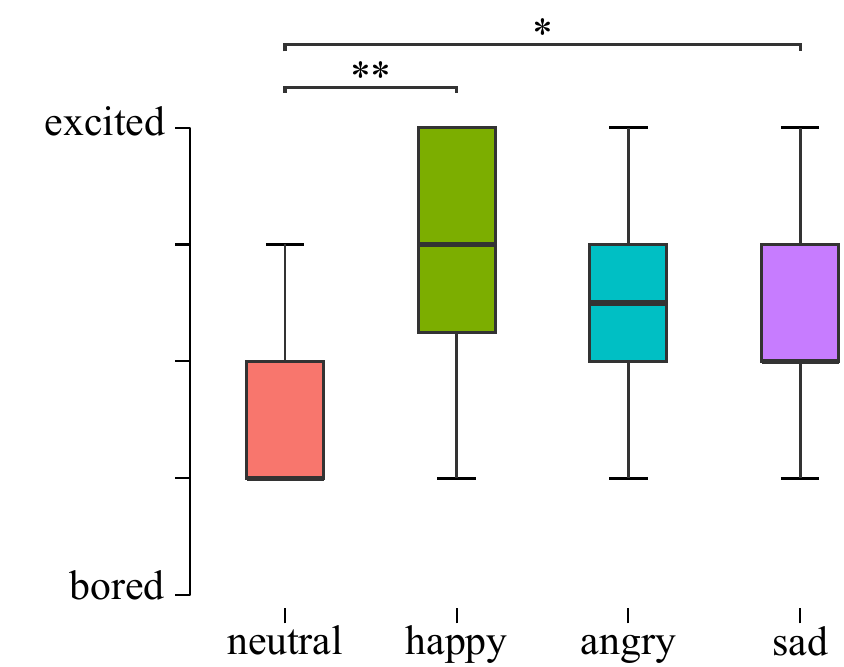}}
    \caption{Perceived emotion between \emph{relaxed} and \emph{stressed} on the left and between \emph{bored} and \emph{excited} on the right. Significant effects are indicated as * for $p_{\text{bonf}}\leq0.05$, ** for $p_{\text{bonf}}\leq0.01$.}
    \label{fig:stress}
    \end{figure}

A repeated measures ANOVA applied to questions regarding participants' feelings of being \emph{relaxed} versus \emph{stressed} and \emph{bored} versus \emph{excited} unveiled a significant main effect for both, with $p < .001$, as visualized in Fig.~\ref{fig:stress}. In both cases, large effect sizes support this finding, with $\eta p^2 = .40$ for perceived stress level and $\eta p^2 = .36$ for perceived boredom.  

Bonferroni post-hoc examinations on stress levels showed that that the \emph{angry} behavior ($M=3.27$, $SE=.31$) caused significantly higher stress levels compared to the \textit{happy} ($M=1.557$, $SE=.18$) and \emph{neutral} behaviors ($M=1.80$, $SE=.24$). The \emph{sad} behavior ($M=2.07$, $SE=.31$) showed lower stress levels than the \emph{angry} behavior but without significance.

Upon evaluating the Bonferroni post-hoc tests for boredom, it showed that the \emph{neutral} behavior ($M=2.64$, $SE=.22$) was predominantly associated with feelings of boredom. It reached significance compared to the \emph{happy} ($M=4.07$, $SE=.26$) and the \emph{sad} behavior ($M=3.29$, $SE=.244$) but not the \emph{angry} behavior ($M=3.50$, $SE=.20$). Furthermore, there is no other measurable difference between the three emotionally expressive behaviors. 

\begin{figure}[t]
    \subfigure{\includegraphics[width=1.55in]{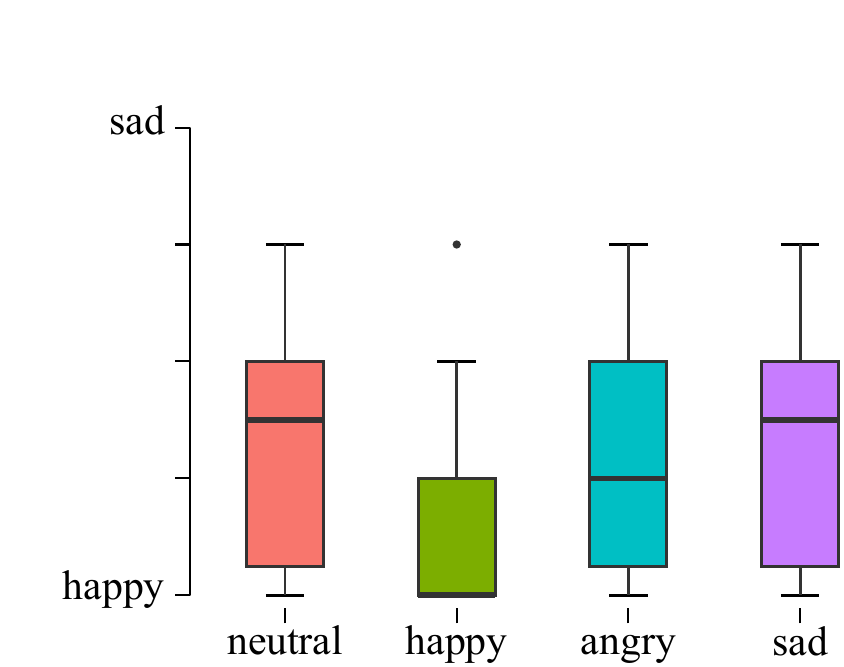}} 
    \subfigure{\includegraphics[width=1.55in]{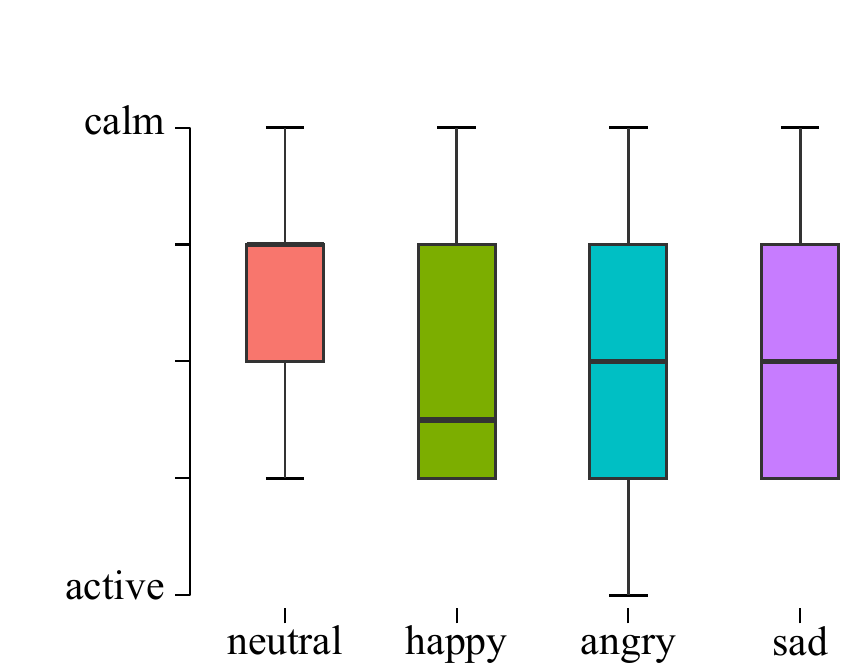}}
    \caption{Perceived emotional valence on the left and arousal on the right.}
    \label{fig:feeling}
\end{figure}

A repeated measures ANOVA was applied to the second category of emotional metrics, representing the participants' emotional valence and arousal. The arousal showed a tendency toward happy in the \emph{happy} condition compared to the other three and the \emph{neutral} resulted in a slight tendency toward calm in the arousal response compared to the other three. However, in the results show no significant main effect in either case, as shown in Fig.~\ref{fig:feeling}. 

\subsection{Time perception}
\begin{figure}[t]
    \subfigure{\includegraphics[width=1.55in]{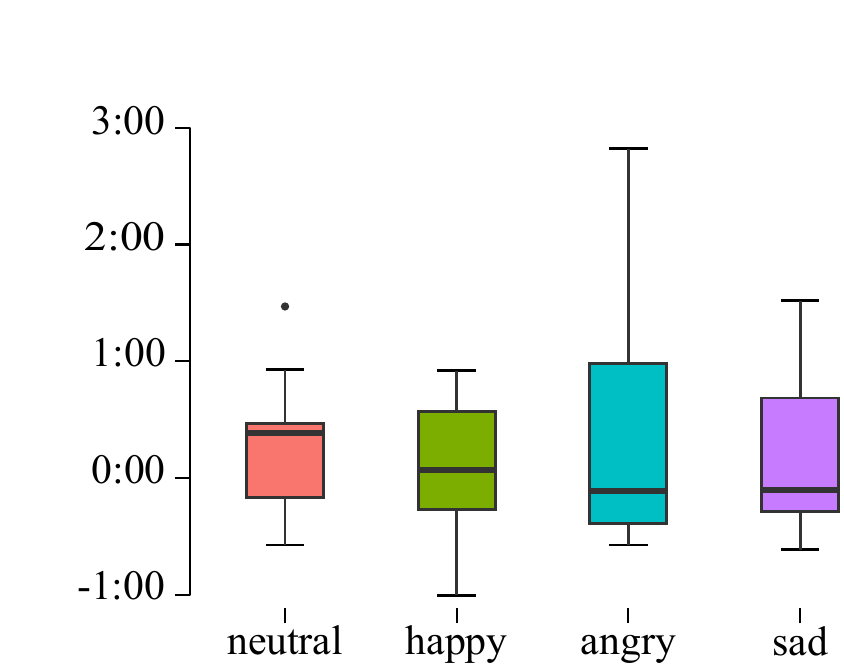}} 
    \subfigure{\includegraphics[width=1.55in]{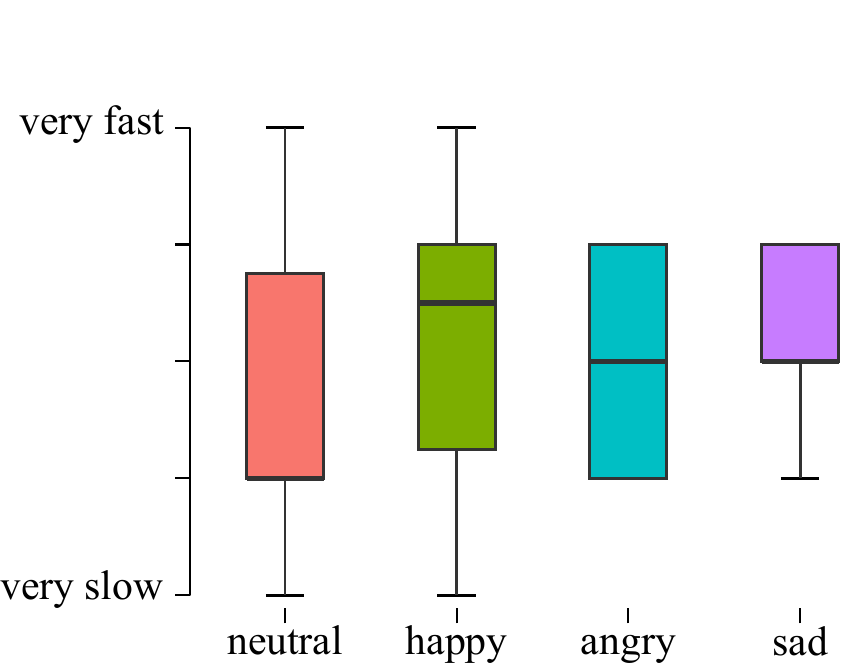}}
    \caption{Duration judgment relative to experiment duration in minutes:seconds on the left and perceived passage of time on the right.}
    \label{fig:time_perception}
\end{figure}

The repeated measures ANOVA results on self-reported data on the passage of time perception and duration judgment relative to the experiment duration show no significant effect on either judgment, as seen in Fig.~\ref{fig:time_perception}. On average, the participants overestimated the duration in all conditions and perceived the time to be passing faster than normal.

\subsection{Behavior classification}
The behavior classification results can be viewed in Table~\ref{tab:class_results}. The \emph{angry} behavior was predominantly recognized, with 93\% of participants correctly identifying it. The \emph{happy} behavior resulted in 57\% of the participants classifying it as such, while another 29\% perceived it as \emph{surprise}. The \emph{neutral} behavior, which was implemented without a specific emotional expression, reflected its intention in the diverse responses of the participants, with answers spread between 7\% and 29\% in all categories. Lastly, the \emph{sad} behavior's representation achieved a comparatively low correct classification score with only 29\% of participants identifying it.

\begin{table}[t]
    \caption{Emotion Classification results}
    \label{tab:class_results}
    \centering
    \begin{tabular}{l | c c c c}
    \toprule
        \textbf{Answer} & \textbf{angry} & \textbf{happy} & \textbf{sad} & \textbf{neutral} \\ 
        \midrule
        Anger       & \underline{93\%}  & 0\%               & 7\%               & 14\% \\
        Happiness   & 7\%               & \underline{57\%}  & 21\%              & \underline{29\%} \\
        Disgust     & 0\%               & 7\%               & 7\%               & 7\% \\
        Fear        & 0\%               & 7\%               & 21\%              & 21\% \\
        Sadness     & 0\%               & 0\%               & \underline{29\%}  & 14\% \\
        Surprise    & 0\%               & 29\%              & 14\%              & 14\% \\
        \bottomrule
    \end{tabular}
\end{table}

\section{Discussion}\label{section:discussion}  
We successfully implemented two distinct emotionally expressive behaviors with \emph{angry} being classified correctly by almost all participants and the \emph{happy} behavior by more than half of the participants. However, the \emph{sad} behavior showed more ambiguity in the self-reported measures. Follow-up responses from the semi-structured interviews reveal that the sinusoidal trajectory, during which the follower lagged further behind, was largely misinterpreted as sadness. Yet, the increased catch-up speed made the \emph{sad} behavior's representation ambiguous, as stated by multiple participants in the semi-structured interviews. Finally, the \emph{neutral} behavior resulted in a desired distribution across all possible answers. These data suggest that while robotic motion can effectively convey emotions, certain emotions (such as anger) might be universally more recognizable than others. The explanation might be that the leader-follower dynamic closely resembles the experience of driving a car, leading many participants to associate the \emph{angry} behavior with the sensation of being tailgated, as mentioned multiple times in the semi-structured interview. Conversely, emotions such as happiness and sadness are rarely conveyed in the context of vehicle navigation, particularly not through sinusoidal movement patterns. This could suggest a unique interpretation of emotional expressions in this specific scenario, where only certain emotions are easily related to the driving experience.

Our results show that emotionally expressive behaviors of a follower robot influence human operators' perception in different ways. Regardless of the behavior type, all three tested implementations shifted the participants' attention toward the follower when compared to the \emph{neutral} follower. This is a feature that could once be leveraged to strategically guide the operator's attention in real-world applications. In situations that demand attention from the following robot, it could switch to an \emph{angry} or \emph{happy} behavior to signal the operator. The \emph{angry} behavior was perceived to keep up better with the leader than the other conditions, while it was the opposite for the \emph{sad} behavior. \emph{Happy} and \emph{neutral} behaviors were perceived equal in this regard. For robotic applications, this could be used to communicate the sense of urgency. This is supported by the increased stress response in the self-reported measures of this behavior and should therefore be used with caution. Finally, the results indicate the highest level of excitement for the \emph{happy} behavior which is underlined with nine out of the 14 participants stating it as their favorite behavior. Also the \emph{sad} behavior resulted in a higher excitement response which could be linked to the ambiguity in participant classification for this behavior. 

Although the study showed clear results on the self-reported measures, no effect was found on emotional valence, arousal, or time perception. Without clear evidence, we suspect that the exposure to the emotionally expressive behaviors might have been too short for the participants to feel a change in their own emotional state. This interpretation is supported by multiple participants highlighting that they enjoyed "playing" with the robots in the experiment and that it was a new experience. The same could also explain the lack of effects on the time perception of the participants.

\section{Conclusion}\label{section:conclusion}  

The evolving field of HRI is moving towards more intuitive and emotionally resonant interactions. With a focus on non-humanoid robots, our study highlights how robot motion can successfully convey emotions, influencing operators' emotional states and their attention. The \emph{angry} behavior, in particular, raised stress levels, suggesting its use for conveying urgency, while the \emph{happy} behavior enhanced engagement and positivity, hinting at ways to improve user experiences. In contrast, the \emph{neutral} behavior demonstrated the potential dullness of interacting with non-expressive robots, and the \emph{sad} behavior's ambiguity highlighted the need for clearer emotional communication through motion.

As robots become more integrated into our daily lives, understanding emotional expressiveness will be important for enhancing interactions, from service robots to collaborative systems. Emotional communication may boost efficiency and decision-making in human-robot collaborations. Our results could be used as starting points for further exploration of the long-term effects of exposure to emotionally expressive robots, the impact of multiple robots, and their motion characteristics on human emotion and behavior. 

Future research could focus on prolonged interactions, diverse users, and complex robot groups to deepen our understanding of emotionally expressive robots and their practical implications. Our findings lay the groundwork for robots that interact with humans more naturally and effectively, paving the way for their application in multi-robot systems.



\bibliography{references.bib}
\bibliographystyle{IEEEtran}

\end{document}